\documentclass{article}
\usepackage{spconf,amsmath,graphicx}
\usepackage{amssymb}
\usepackage{algorithmic}
\usepackage{array}
\usepackage{subcaption}
\usepackage{textcomp}
\usepackage{stfloats}
\usepackage{url}
\usepackage{verbatim}
\usepackage{graphicx}
\usepackage{hyperref}
\hypersetup{hidelinks}
\usepackage{soul}

\usepackage{comment}
\usepackage{booktabs}
\usepackage{multirow}
\usepackage[flushleft]{threeparttable}
\usepackage{makecell}
\usepackage[square,numbers,sort&compress]{natbib}
\usepackage{bm}
\usepackage{ulem}
\makeatletter
\newcommand\figcaption{\def\@captype{figure}\caption}
\newcommand\tabcaption{\def\@captype{table}\caption}
\makeatother

\usepackage{xcolor,sectsty}

\title{
 MMFormer: Multimodal Transformer Using Multiscale Self-Attention for Remote Sensing Image Classification
}
\name{\makecell{Bo~Zhang$^{1}$, Zuheng~Ming$^{2}$, Wei~Feng$^{3}$, Yaqian~Liu$^{1}$, Liang~He$^{1\ast}$\thanks{$\ast$Corresponding author. 2021050018@nwpu.edu.cn(Liang~He)}, Kaixing~Zhao$^{1\ast}$\thanks{$\ast$Corresponding author. kaixing.zhao@nwpu.edu.cn(Kaixing~Zhao)}}
}
\address{$^{1}$~School of Software, Northwestern Polytechnical University\\ $^{2}$~Laboratoire L2TI, Institut Galilée, Université Sorbonne Paris Nord\\ $^{3}$~School of Electronic Engineering, Xidian University
}

\begin{document}
\maketitle
\begin{abstract} 
To benefit the complementary information between heterogeneous data,
we introduce a new Multimodal Transformer (MMFormer) for Remote Sensing (RS) image classification using Hyperspectral Image (HSI) accompanied by another source of data such as Light Detection and Ranging (LiDAR). Compared with traditional Vision Transformer (ViT) lacking inductive biases of convolutions, 
we first introduce convolutional layers to our MMFormer to tokenize patches from multimodal data of HSI and LiDAR. Then we propose a Multi-scale Multi-head Self-Attention (MSMHSA) module to address the problem of compatibility which often limits to fuse HSI with high spectral resolution and LiDAR with relatively low spatial resolution. The proposed MSMHSA module can incorporate HSI to LiDAR data in a coarse-to-fine manner enabling us to learn a fine-grained representation.
Extensive experiments on widely used benchmarks (e.g., Trento and MUUFL) 
demonstrate the effectiveness and superiority of our proposed MMFormer for RS image classification.
\end{abstract}
\section{Introduction}
\label{sec:intro}
Remote sensing (RS) has been playing an vital role in various Earth Observing (EO) tasks because of its rapid imaging feature and wide application prospect. Generally, RS can be used (but not limited) in different tasks, such as landcover classification \cite{Ahmad_2022,bartholome2005glc2000,roy2021revisiting}, mineral and forest resources exploration \cite{koetz2008multi}, object/target detection \cite{wu2019orsim, wu2019fourier}, environmental monitoring \cite{ustin2004manual}, urban planning \cite{chen2020classification}, biodiversity conservation, as well as disaster response and management. With the ever-growing availability of RS data, research on RS has begun to shift to data-driven methods, and various Machine Learning (ML) and Deep Learning (DL) models have been applied in RS systems. However, in the past few years, most studies focused only on single EO sensors, such as HSI sensors, rather than combining different types of sensor data.
Although HSI acquired from different sensors can provide more rich spectral information, it cannot differentiate the landcover objects such as roads and roofs using the same materials~\cite{PedramGhamisi2015LandcoverCU}. In contrast, LiDAR provides elevation information that allows distinguishing objects with identical spectral signatures but different elevations, such as roads and roof built-in cement~\cite{MFT}. Multimodal data integration for RS classification can a solution to the dilemma.

In recent years, DL methods have been widely used in multimodal data fusion for RS image classification~\cite{KonstantinosMakantasis2015DeepSL,AminaBenHamida20183DDL}. More recently, Vision Transformers (ViT)~\cite{vaswani2017attention} using self-attention considering the local-global context of an image or image sequence has been becoming a new mainstream for RS image classification. SpectralFormer~\cite{DanfengHong2021SpectralFormerRH} learned the spectral representation of neighboring bands using a cross-layer encoder module of ViT. However, it neglected spatial information and only uses spectral information. Swalpa et al. proposed MFT~\cite{MFT} which incorporates HSI and other source data to construct a multimodal ViT for RS image classification. Nevertheless, MFT did not consider the compatibility problem when fusing the data with a large resolution gap such as HSI and LiDAR data which may affect the landcover classification in complex scenes~\cite{LianruGao2021SpectralSO}. To leverage the complementary information between the different modalities, we propose a multimodal transformer using self-attention between two modalities HSI and LiDAR data for RS image classification. Instead of using a vanilla ViT as in MFT or SpectralForm, we propose a Multi-scale Multi-head Self-Attention (MSMHSA) using different heads of transformer aiming to better fuse the feature representations with different resolutions of heterogeneous data from different modalities. Besides, we also introduce convolutions to our MMFormer using Convolutional Tokenization to tokenize HSI and LiDAR rather than using linear projection to encode Q, K, V feature maps for self-attention.

The main contributions of this paper can be summarized as follows: 1) We design a Multimodal Transformer (MMFormer) for RS image classification using HSI and LiDAR data. 2) A Multi-scale Multi-Head Self-Attention (MSMHSA) implemented on a single transformer allows better fusing of multimodal data of different resolutions. 3) The introduction of convolutions to MMFormer to integrate desirable convolution proprieties. 4) The state-of-art performance demonstrates that the proposed MMFormer can serve as an effective new backbone for multimodal RS image classification.

\section{Related works}
In the past few years, various traditional methods have been investigated to extract more effective information from multi-source RS data, such as random forest (RF), morphological profiles (MPs) \cite{benediktsson2005classification}, attribute profiles (APs) \cite{dalla2010morphological}, extinction profiles (EPs) \cite{ghamisi2016extinction} and subspace learning~\cite{de2003framework}. Ham et al. investigated hierarchical classifiers based on RF to improve the generalization in analysis of quantity limited hyperspectral data\cite{Ham2004RandomForest}. Recently, deep learning techniques have been widely used in the task of multimodal data fusion for RS data classification and have shown excellent feature extraction abilities. In~\cite{KonstantinosMakantasis2015DeepSL}, Makantasis et al. exploited a CNN2D based network to encode pixels' spectral and spatial information. Vision Transformers methods used self-attention to learn the local-global context of an HSI data. SpectralFormer~\cite{DanfengHong2021SpectralFormerRH} used a cross-layer encoder module of ViT to learn the spectral features between HSI bands. And Swalpa et al. proposed MFT~\cite{MFT} which leverage the complementary information between the different modalities to construct a multimodal ViT for RS image classification. Inspired by the former methods, we propose a Multi-scale Multi-head Self-Attention transformer to improve the overall fusion and classification accuracy.

\section{Methodology}
\label{sec:Method}

\subsection{Overall Architecture}
\label{sec:Architecture}

\begin{figure*}[t]
 \vspace{-0.8em}
\centering
  \includegraphics[scale=0.45]{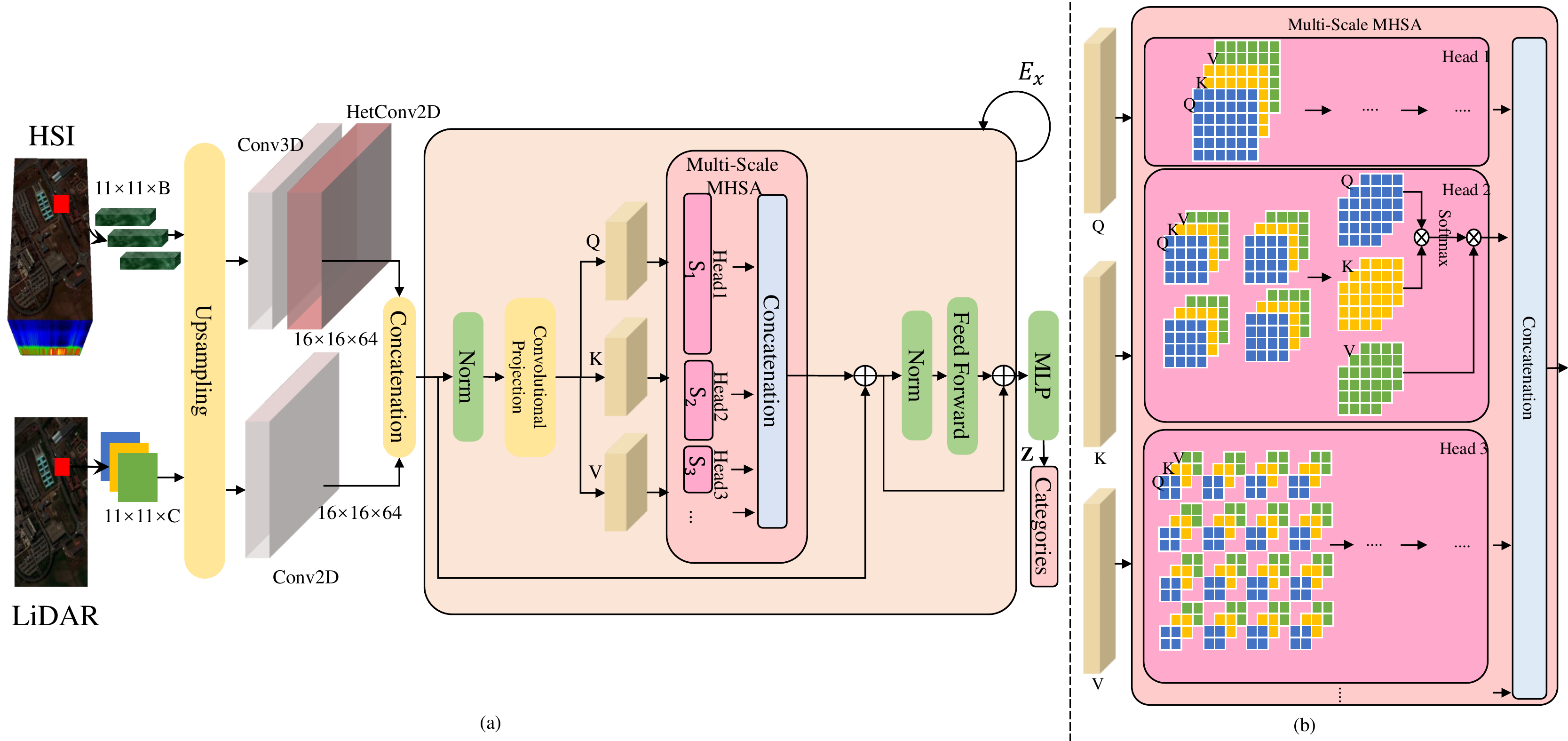}
  \vspace{-0.8em}
  \caption{(a) Overall architecture of the proposed : Multimodal Transformer Using Multiscale Self-Attention for RS image classification (MMFormer) introducing convolutions to transformers. (b) The proposed Multi-scale Multi-Head Self-Attention (MSMHSA) module implemented on a single transformer.}
  \label{fig:MMFormer}
  \vspace{-0.8em}
\end{figure*}

An overview of the proposed MMFormer is depicted in ~\figureautorefname~\ref{fig:MMFormer}. 
To balance the performance and the parameters of model, we set the proposed transformer depth to 2. \\


\vspace{0.3em}
\noindent\textbf{\textit{Convolutional Tokenization}}.\quad Unlike the vanilla Transformer, we use convolutional embedding instead of linear projection to tokenize $\mathbf{Q}/\mathbf{K}/\mathbf{V}\in\mathbb{R}^{ C_A\times H_A \times W_A}$ feature maps. Firstly, the 11*11 data cubes are padding to 16*16. And sequential layers Conv3D~\cite{AminaBenHamida20183DDL} and HetConv2D~\cite{hetconv2019} are used to extract the feature map of HSI cube and reduce the HSI spectral channels down to 64. A Conv2D layer is used to extract LiDAR cube's feature map and expend the band to 64. After that, the feature maps, concatenated in the last dimension, act as the input of convolutional embedding to get the $\mathbf{Q}$, $\mathbf{K}$ and $\mathbf{V}$. \par 
Then the obtained $\mathbf{Q}$, $\mathbf{K}$ and $\mathbf{V}$ are fed into the MSMHSA module to learn the local-global dependencies in the fusion data. In the output layer of MSMHSA module, we also replace the linear layer and layer normalization with a convolution layer(with kernel size = 3, padding = 1) and a LeakyReLU activation function. The mentioned modules could be defined as follow:
\begin{equation}
    \mathbf{Q,K,V} = Conv2D(\mathbf{X_{in}}, k = (1,1))
\end{equation}
\begin{equation}
\left\{\begin{aligned}
X_{out} &= Conv2D(X_{in}, k = (3,3), p = (1,1)) \\
      X &= LeakyReLU(\mathbf{X_{out}}, 0.2) \\
\end{aligned}\right.
\end{equation}
The details of MSMHSA module are described in Section~\ref{sec:MSMHSA}. Finally, we add Feed-Forward Network (FFN) with Norm layers at the end of transformer. In this work, linear projection layers in FFN are also replaced by convolution layers. As in ViT~\cite{dosovitskiy2020image}, an MLP Head is connected to the transformer to generate the classification embeddings.

\subsection{Multi-scale Multi-Head Self-Attention (MSMHSA)}
\label{sec:MSMHSA}
The goal of MSMHSA, as shown in \figureautorefname~\ref{fig:MMFormer}(b), is to introduce a pyramid structure into the self-attention module to generate multi-scale feature maps which can employ pixel-level fine-grained feature fusion on the complimentary features. The proposed MSMHSA is applied on different heads of each layer of transformer. All the heads share the similar protocol to calculate the self-attention. \par
In particular, the feature maps $\mathbf{Q/K/V}$ are equally divided to each head along the dimension before inputting them to MSMHSA. \par
Given a transformer with three heads fed by input feature maps $\mathbf{Q/K/V}$ of size $C_A \times H_A \times W_A$, the feature maps for each head are $\mathbf{Q_i/K_i/V_i}\in \mathbb{R}^{\frac{C_A}{3}\times H_A \times W_A} $. For the first head $Head_1$, we take a full-size patch $\mathbf{q_1/k_1/v_1 }\in\mathbb{R}^{\frac{C_A}{3} \times H_A \times W_A}$ to calculate a global self-attention feature map for the first head $Head_1$. We can obtain the self-attention feature map $\mathbf{h_1}\in\mathbb{R}^{\frac{C_A}{3} \times H_A \times W_A}$ of $Head_1$. Then for $Head_2$, we divide $\mathbf{Q_2/K_2/V_2}$ into $2^2$ patches, each patch $\mathbf{q_2/k_2/v_2}$ of size ${4 \times \frac{C_A}{3} \times \frac{H_A}{2} \times \frac{W_A}{2}}$ and then obtain $\mathbf{h_2}\in\mathbb{R}^{4 \times \frac{C_A}{3} \times \frac{H_A}{2} \times \frac{W_A}{2}}$ of $Head_2$. We continue to divide $\mathbf{Q_3/K_3/V_3}$ into $4^2$ patches to calculate the self-attention feature map $\mathbf{h_3}\in\mathbb{R}^{16\times \frac{C_A}{3} \times \frac{H_A}{4} \times \frac{W_A}{4}}$ of $Head_3$. Finally, we concatenate the obtained self-attention feature maps $\{\mathbf{h_1},\mathbf{h_2},\mathbf{h_3}\}$ to generate the final multi-scale attention feature map $\mathbf{H}\in\mathbb{R}^{C_A \times H_A \times W_A}$ in layer $L_i$ (We need to reshape $\mathbf{h_2},\mathbf{h_3}$ to be consistent with $\mathbf{h_1}$):
\begin{equation}
\mathbf{H} = \mathbf{Concat(\mathbf{h_1}, \mathbf{Reshape(h_2)}, \mathbf{Reshape(h_3)})}.
\end{equation}
\vspace{-0.8em}
The self-attention feature map $\mathbf{h_i}$ is given by:
\begin{equation}
\label{MSMHSA}
\mathbf{h_i} = \Sigma_m^N\Sigma_n^N{\mathbf{Softmax}(\frac{\mathbf{q_{i,m}k_{i,n}^T}}{\sqrt{d_{head_{i,n}}}})\mathbf{v_{i,n}}},
\end{equation}
where $i$ is corresponding to the $i$th head $Head_i$, $\mathbf{q_{i,m}}$ is the $m$th patch partitioned from feature map $\mathbf{Q_i}$ for the $i$th head $Head_i$, $\mathbf{k_{i,n}/v_{i,n}}$ are the $n$th patches partitioned from feature maps $\mathbf{K_i/V_i}$ for the $i$th head $Head_i$. Then, $\mathbf{q_{i,m}/k_{i,n}/v_{i,n} }\in\mathbb{R}^{\frac{C_A}{3} \times \frac{H_A}{l} \times \frac{W_A}{l}}, l\in[1,2,4]$. and $N$ is the total number of patches of $i$th head, i.e., $N=T\times l^2, l\in[1,2,4]$. $d_{head_{ij}}$ is the dimension of the $\mathbf{q_{i,m}}$. For each head $Head_i$, we stack the partitioned patches $\mathbf{q_{i}/k_{i}/v_{i}}$ from all frames to calculate the self-attention feature map $\mathbf{h_i}$. Thus, the self-attention of each head always considers simultaneously the local attention focusing on local spatial information, i.e., using $\mathbf{q_{i,m}/k_{i,n}}$ from the neighborhood to calculate self-attention, and global attention focusing on the global context information calculated by $\mathbf{q_{i,m}/k_{i,n}}$ from far regions within the image. Instead of adding an extra token as ViT for image classification, we learn a representation of a data sequence, i.e., the output of MLP Head $\mathbf{Z}$, based on all tokens with cross-entropy loss to conduct final landcover classification in this work.

\section{EXPERIMENTS AND ANALYSIS}
\label{sec:experimentsAndAnalysis}

\subsection{Experimental Setup}
\textbf{(1) Data Description}
To test the effectiveness of our proposed MMFormer, we conduct experiments on two widely used hyperspectral and LiDAR fusion datasets.
\vspace{0.3em}\\
\noindent\textbf{\textit{Trento dataset}}:\quad this dataset contains similarly one HSI and one LiDAR data, collected from a rural area south of the city of Trento, Italy. HSI and LiDAR data are collected with 63 bands and 1 band specifically. Both types of data contain 166 × 600 pixels with a spatial resolution of 1 m, containing a total of 6 different classes.
\vspace{0.3em}\\
\noindent\textbf{\textit{MUUFL dataset}}:\quad this dataset was collected over the University of Southern Mississippi Gulf Park, both the HSI and LiDAR data contain 325 × 220 pixels. HSI initially contained 72 bands, however initial and final four bands are removed due to noise issues, the remaining 64 bands were used for the experiment, and the LiDAR data contained 2 bands. There are 11 different classes.

\begin{figure*}[t]
 \vspace{-0.8em}
\centering
  \includegraphics[scale=0.45]{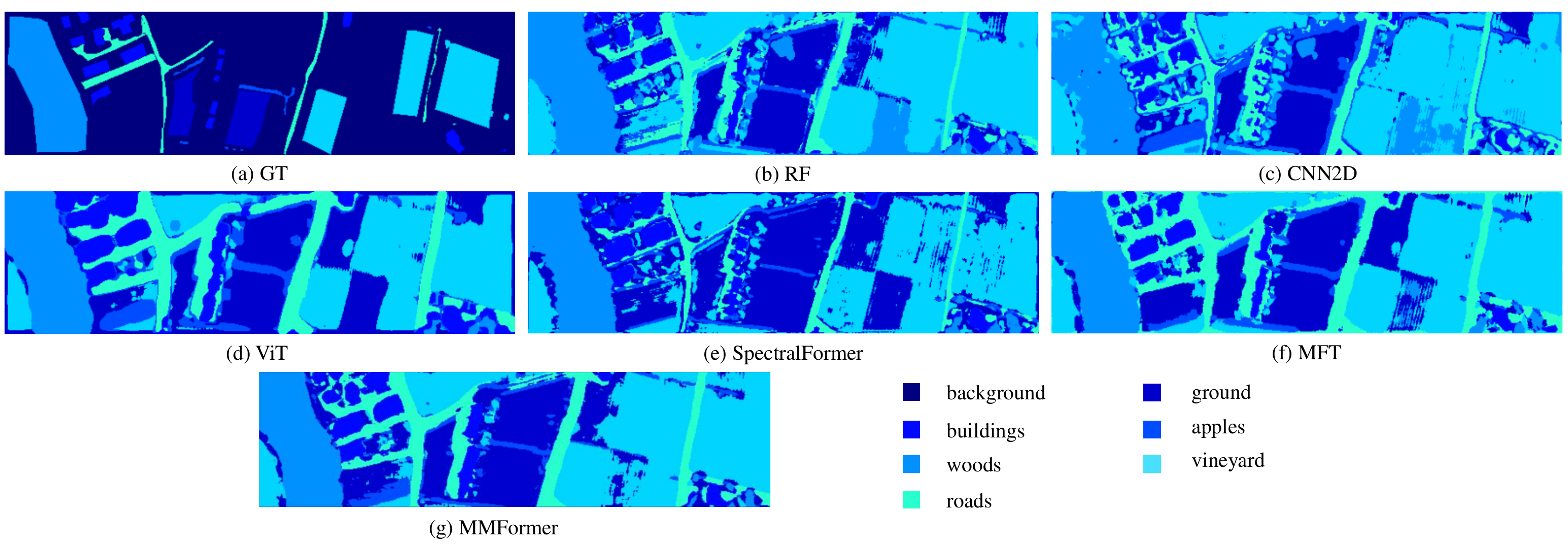}
  \vspace{-0.8em}
  \caption{(a) Ground truth and classiﬁcation maps inferred by: (b) RF,(c) CNN2D, (d) ViT, (e) SpectralFormer~\cite{DanfengHong2021SpectralFormerRH}, (f) MFT, (g) MMFormer on HSI and LiDAR data for the Trento dataset}
  \label{fig:MMFormer Visualization}
  \vspace{-0.8em}
\end{figure*}

\noindent\textbf{(2) Experimental Setup}
All the tests have been performed on a CentOS Linux server(release 7.9.2009) and a single GPU, Nvidia 3090 with 24576MB of VRAM.\par
A batch size of 64 and 500 has been used for training and testing the performance of the considered models. The models have been trained with a Adam optimizer (learning rate = 5e-4, weight decay = 5e-3) and a step scheduler (step size = 50 and gamma = 0.9). Each experiment has been conducted using 500 epochs repeating 3 times and the average and standard deviations are reported. The source code was implemented using PyTorch 1.12.1 and Python 3.8.7.

\noindent\textbf{(3) Evaluation metrics}
We evaluate the classification performance of our model quantitatively using three widely-used metrics, i.e., overall accuracy (OA), average accuracy (AA) and statistical Kappa ($\mathbf{\kappa}$) coefficients.

\begin{table}[]\small
\centering
\vspace{-1.0em}
\caption{OA, AA and Kappa values on the Trento data (in \%) by considering HSI and LiDAR data. The Best is shown in \textbf{bold}.}
\vspace{-0.8em}
\scalebox{0.6}{
    \begin{tabular}{c||c | c|c|c|c|c}
        \hline
        Class & \multirow{2}{*}{RF } & \multirow{2}{*}{CNN2D } & \multirow{2}{*}{ViT } & Spectral-& \multirow{2}{*}{MFT}           & \multirow{2}{*}{MMFormer} \\
        No.&&&&Former&& \\
        \hline
        1 & 83.73 ± 00.06 & 96.98 ± 00.21 & 90.87 ± 00.77 & 96.76 ± 01.71 & 98.23 ± 00.38 & \textbf{99.71 ± 0.25}   \\
        2 & 96.30 ± 00.06 & 97.56 ± 00.14 & 99.32 ± 00.77 & 97.25 ± 00.66 & \textbf{99.34 ± 00.02} & 98.06 ± 0.80   \\
        3 & 70.94 ± 01.55 & 55.35 ± 00.00 & 92.69 ± 01.53 & 58.47 ± 11.54 & 89.84 ± 09.00 & \textbf{94.47 ± 1.77}   \\
        4 & 99.73 ± 00.07 & 99.66 ± 00.03 & \textbf{100.0 ± 00.00} & 99.24 ± 00.21 & 99.82 ± 00.26 & 99.96 ± 0.02   \\
        5 & 95.35 ± 00.25 & 99.56 ± 00.07 & 97.77 ± 00.86 & 93.52 ± 01.75 & \textbf{99.93 ± 00.05} & 99.90 ± 0.07   \\
        6 & 72.63 ± 00.90 & 76.91 ± 00.15 & 86.72 ± 02.02 & 73.39 ± 06.78 & 88.72 ± 00.94 & \textbf{95.34 ± 1.32}   \\
        \hline\hline
       OA & 92.57 ± 00.07 & 96.14 ± 00.03 & 96.47 ± 00.49 & 93.51 ± 01.27 & 98.32 ± 00.25 & \textbf{99.18 ± 0.02}   \\
       AA & 86.45 ± 00.32 & 87.67 ± 00.04 & 94.56 ± 00.57 & 86.44 ± 02.96 & 95.98 ± 01.64 & \textbf{97.91 ± 0.25}   \\
 $\kappa$ & 90.11 ± 00.09 & 94.83 ± 00.04 & 95.28 ± 00.65 & 91.36 ± 01.67 & 97.75 ± 00.00 & \textbf{98.90 ± 0.02}   \\

        \bottomrule[1pt]
    \end{tabular}}
\label{tab:TrentoResult}
\end{table}

\begin{table}[]\small
\centering
\vspace{-1.0em}
\caption{OA, AA and Kappa values on the MUUFL data (in \%) by considering HSI and LiDAR data. The Best is shown in \textbf{bold}.}
\vspace{-0.8em}
\scalebox{0.6}{
    \begin{tabular}{c||c | c|c|c|c|c}
        \hline
        Class & \multirow{2}{*}{RF } & \multirow{2}{*}{CNN2D } & \multirow{2}{*}{ViT } & Spectral-& \multirow{2}{*}{MFT}           & \multirow{2}{*}{MMFormer} \\
        No.&&&&Former&& \\
        \hline
        1 & 95.42 ± 00.09 & 95.79 ± 00.11& 97.85 ± 00.29 & 97.30 ± 00.83 & 97.90 ± 00.39 & \textbf{98.88 ± 0.15} \\
        2 & 74.03 ± 00.11 & 72.76 ± 00.58& 76.06 ± 02.40 & 69.35 ± 05.16 & \textbf{92.11 ± 01.58} & 88.84 ± 1.66 \\
        3 & 75.81 ± 00.38 & 78.92 ± 00.52& 87.58 ± 03.46 & 78.48 ± 03.41 & \textbf{91.80 ± 00.82} & 90.00 ± 0.80 \\
        4 & 68.59 ± 00.77 & 83.59 ± 00.99& 92.05 ± 02.31 & 82.63 ± 03.68 & 91.59 ± 02.25 & \textbf{95.19 ± 0.24} \\
        5 & 88.17 ± 00.18 & 78.29 ± 01.12& 94.73 ± 00.60 & 87.91 ± 02.97 & \textbf{95.60 ± 01.21} & 95.28 ± 0.48 \\
        6 & 77.28 ± 00.93 & 50.34 ± 02.13& 82.02 ± 01.13 & 58.77 ± 02.76 & 88.19 ± 03.49 & \textbf{88.48 ± 0.97} \\
        7 & 64.83 ± 00.97 & 79.70 ± 00.26& 87.11 ± 01.54 & 85.87 ± 00.62 & 90.27 ± 02.13 & \textbf{92.94 ± 1.14} \\
        8 & 93.29 ± 00.27 & 71.95 ± 01.10& 97.60 ± 00.16 & 95.60 ± 01.26 & 97.26 ± 00.53 & \textbf{97.84 ± 0.53} \\
        9 & 19.15 ± 01.37 & 43.92 ± 01.24& 57.83 ± 04.45 & 53.52 ± 04.32 & 61.35 ± 03.80 & \textbf{65.02 ± 1.79} \\
       10 & 04.41 ± 00.72 & 12.45 ± 00.27& 31.99 ± 08.86 & 08.43 ± 02.22 & 17.43 ± 04.63 & \textbf{36.97 ± 3.39} \\
       11 & 71.88 ± 00.84 & 26.82 ± 02.60& 58.72 ± 03.85 & 35.29 ± 06.00 & 72.79 ± 09.25 & \textbf{80.85 ± 5.58} \\
        \hline\hline
       OA & 85.32 ± 00.09 & 83.40 ± 00.04& 92.15 ± 00.19 & 88.25 ± 00.56 & 94.34 ± 00.07 & \textbf{94.73 ± 0.20} \\
       AA & 66.62 ± 00.16 & 63.14 ± 00.21& 78.50 ± 01.28 & 68.47 ± 01.44 & 81.48 ± 00.70 & \textbf{84.57 ± 0.35} \\
 $\kappa$ & 80.39 ± 00.12 & 77.94 ± 00.06& 89.56 ± 00.27 & 84.40 ± 00.77 & 92.51 ± 00.10 & \textbf{93.02 ± 0.26} \\

        \bottomrule[1pt]
    \end{tabular}}
\label{tab:MUUFLResult}
\end{table}

\subsection{Quantitative analysis}
\tableautorefname~\ref{tab:TrentoResult} and \tableautorefname~\ref{tab:MUUFLResult} report the quantitative OA, AA, $kappa$ and each class accuracy on two widely used datasets Trento and MUUFL to compare the proposed MMFormer with other state-of-art methods, i.e., RF~\cite{Ham2004RandomForest}, CNN2D~\cite{KonstantinosMakantasis2015DeepSL}, ViT~\cite{dosovitskiy2020image}, SpectralFormer~\cite{DanfengHong2021SpectralFormerRH} and MFT~\cite{MFT}. Our model MMFormer obtains the best  performance on all three indices on both two benchmarks, such as 99.18\% and 94.73\% OA, 97.91\% and 84.57\% AA and 98.90\% and 93.02\% Kappa 
on Trento and MUUFL datasets. Our model has also achieved the best  performance for almost each class accuracy, and even it has gained 20\% improvement on landcover Yellow-Curb (class 10 ) compared to the latest MFT on MUUFL data.

\subsection{Visualization}
~\figureautorefname~\ref{fig:MMFormer Visualization} illustrates a qualitative evaluation by visualizing the classification maps obtained by different methods Trento dataset using HSI and LiDAR data. The MSMHSA module helps to learn the features in a fine-to-coarse manner, which obtains a classification map with less noise and finer details.

\subsection{Ablation study}
All ablation studies are conducted on the Trento dataset. And the best results in the tables are shown in bold.
\vspace{0.3em}\\
\noindent\textbf{Effectiveness of multimodal fusion.}\quad
In order to evaluate the effectiveness of multimodal fusion, we trained the proposed model respectively on single modality and multimodalities. Only one branch has been used to input the data when training the model in single modality as shown in ~\figureautorefname~\ref{fig:MMFormer}. In~\tableautorefname~\ref{tab:multimodalAblation}, we can see that the performance of multimodal model is superior to single-modal either HSI or LiDAR for all three indices, i.e., OA, AA, $\kappa$ and almost for each class accuracy, which demonstrate the effectiveness of our multimodal transformer MMFormer for RS image classification.
\vspace{0.3em}\\
\noindent\textbf{Effectiveness of multi-scale MHSA.}\quad
In~\tableautorefname~\ref{tab:MSMHSA}, we can see that the model using MSMHSA performs always better than the one using single-scale MHSA, e.g., the model using 16$\times$16, 4$\times$4, 2$\times$2 gains $1\%$ improvement in terms  OA,AA,$\kappa$.

\begin{figure}
\begin{minipage}{0.23\textwidth}
\tabcaption{Multimodal v.s. single-modal MMFormer.}
\centering
\vspace{-0.8em}
\scalebox{0.45}{
    \begin{tabular}{c||c | c|c}
        \hline
        Class & LiDAR only & HSI only & Multimodal \\
        \hline
        1 & 97.93 ± 0.56 & 99.21 ± 0.27 & \textbf{99.71 ± 0.25} \\
        2 & 74.47 ± 4.60 & 91.20 ± 2.48 & \textbf{98.06 ± 0.80} \\
        3 & 58.11 ± 5.98 & 93.58 ± 2.18 & \textbf{94.47 ± 1.77} \\
        4 & 93.79 ± 0.43 & 99.83 ± 0.20 & \textbf{99.96 ± 0.02} \\
        5 & 96.67 ± 0.56 & \textbf{99.96 ± 0.03} & 99.90 ± 0.07 \\
        6 & 67.03 ± 3.18 & 77.34 ± 4.60 & \textbf{95.34 ± 1.32} \\
        \hline\hline
       OA & 90.29 ± 0.45 & 96.56 ± 0.60 & \textbf{99.18 ± 0.02} \\
       AA & 81.33 ± 1.47 & 93.52 ± 1.35 & \textbf{97.91 ± 0.25} \\
 $\kappa$ & 86.99 ± 0.61 & 95.39 ± 0.81 & \textbf{98.90 ± 0.02} \\
        \bottomrule[1pt]
    \end{tabular}
 }
\vspace{3.34em}
\label{tab:multimodalAblation}
\end{minipage}
\begin{minipage}{0.23\textwidth}
\tabcaption{Effectiveness of multiscale MMFormer.}
\centering
\vspace{-0.8em}
\scalebox{0.45}{
\begin{tabular}{ccccc|c|c}
        \hline
        16*16 & 8*8 & 4*4 & 2*2 & OA & AA & $\kappa$\\
        \hline
        $\surd$&        &        &        & 99.04 & 97.29 & 98.71 \\
               & $\surd$&        &        & 98.46 & 97.03 & 97.94 \\
               &        & $\surd$&        & 98.89 & 97.50 & 98.51 \\
               &        &        & $\surd$& 98.97 & 96.93 & 98.62 \\
        $\surd$& $\surd$&        &        & 99.06 & 96.72 & 98.74 \\
        $\surd$&        & $\surd$&        & 99.18 & 97.61 & 98.90 \\
        $\surd$&        &        & $\surd$& 99.17 & 97.58 & 98.89 \\
               & $\surd$& $\surd$&        & 98.78 & 96.96 & 98.36 \\
               & $\surd$&        & $\surd$& 98.70 & 96.11 & 98.26 \\
               &        & $\surd$& $\surd$& 99.06 & 98.09 & 98.74 \\
        $\surd$& $\surd$& $\surd$&        & 99.13 & 97.21 & 98.83 \\
        $\surd$& $\surd$&        & $\surd$& 99.05 & 97.42 & 98.73 \\
   \uline{\bm{$\surd$}}&        & \uline{\bm{$\surd$}}& \uline{\bm{$\surd$}}& \textbf{99.18} & \textbf{97.91} & \textbf{98.91} \\
               & $\surd$& $\surd$& $\surd$& 98.71 & 97.28 & 98.27 \\
        $\surd$& $\surd$& $\surd$& $\surd$& 99.14 & 97.22 & 98.85 \\

        \bottomrule[1pt]
    \end{tabular}
}
\label{tab:MSMHSA}
\end{minipage}
\end{figure}

\section{Conclusion}
\vspace{-0.5em}
We design a Multimodal Transformer (MMFormer) which enables us to explore the complementary information between spectral information in HSI and spatial information in LiDAR for RS image classification. The proposed Multi-scale Multi-Head Self-Attention (MSMHSA) in Transformer aims to better fuse the multimodal data with very different resolutions. We also introduce convolutions to our model to integrate desirable proprieties of CNNs which can gain a good computation-accuracy balance. The effectiveness and state-of-art performance demonstrate that the proposed MMFormer can serve as a new valuable backbone for RS image classification.

\vfill\pagebreak
\bibliographystyle{IEEEbib}
\bibliography{strings}

\end{document}